\documentclass{article}
\usepackage{ijcai26}

\usepackage{times}
\usepackage{soul}
\usepackage{url}
\usepackage[hidelinks]{hyperref}
\usepackage[utf8]{inputenc}
\usepackage[small]{caption}
\usepackage{graphicx}
\usepackage{amsmath}
\usepackage{amsthm}
\usepackage{amssymb}
\usepackage{booktabs}
\usepackage{algorithm}
\usepackage{algorithmic}
\usepackage[switch]{lineno}

\title{ILR-SMO: Iterative Latent Refinement for Robust Spatial Multi-Omics Integration}

\author{
Anqi Yu$^1$
\and
Xudong Xu$^1$
\and
Jianzhi Lu$^1$
\and
Yuqi Sun$^1$
\and
Bingguo Chen$^1$
\and
Chenxi Ma$^{1,*}$
\and
Weimin Tan$^{1,*}$
\and
Bo Yan$^{1,*}$\\
\affiliations
$^1$College of Computer Science and Artificial Intelligence, Shanghai Key Laboratory of Intelligent Information Processing, Fudan University, Shanghai 200438, China.\\
\emails
24110240106@m.fudan.edu.cn,
25113050293@m.fudan.edu.cn,
24110240058@m.fudan.edu.cn,
22110240034@m.fudan.edu.cn,
25213050109@m.fudan.edu.cn,
cxma17@fudan.edu.cn,
wmtan@fudan.edu.cn,
byan@fudan.edu.cn
}

\begin{document}



\maketitle

\begingroup
\renewcommand{\thefootnote}{\fnsymbol{footnote}}
\footnotetext[1]{Corresponding authors.}
\endgroup

\begingroup
\renewcommand{\thefootnote}{}
\footnotetext[0]{%
\textcopyright~2026 International Joint Conferences on
Artificial Intelligence Organization
(\href{https://www.ijcai.org/}{IJCAI}).
This is an author-corrected version of the paper published
in the Proceedings of IJCAI-ECAI 2026.
The original publication is available at
\href{https://doi.org/10.24963/ijcai.2026/582}
{doi:10.24963/ijcai.2026/582}.%
}
\endgroup

\begin{abstract}
    Spatial multi-omics technologies jointly profile diverse molecular modalities with spatial context, providing a comprehensive view of cellular heterogeneity and tissue organization. To integrate spatial multi-omics data and identify spatial domains, a wide range of unsupervised methods has been proposed. However, recent approaches rely on single-step fusion, directly aggregating heterogeneous modalities into a shared representation, making embeddings sensitive to modality imbalance and measurement noise that are inherent to spatial multi-omics data, as well as unstable optimization under weak supervision. Here, we propose ILR-SMO (Iterative Latent Refinement for Spatial Multi-Omics), a unified self-supervised framework that reformulates multimodal integration as a stability-aware refinement process over a shared latent representation. Instead of single-step plain fusion, ILR-SMO progressively integrates modality-specific information through sequential updates, allowing information to be injected in a controlled manner. This incremental refinement enables later updates to correct or compensate for earlier deviations, resulting in more stable and robust representation learning under heterogeneous and noisy modalities. ILR-SMO further incorporates reliability-aware modality gating for adaptive modulation of modality contributions, and employs a joint objective to enforce spatial coherence while preventing representation collapse. Extensive experiments on five spatial multi-omics benchmarks demonstrate that ILR-SMO consistently outperforms seven state-of-the-art methods and exhibits strong robustness across diverse settings. 
\end{abstract}

\section{Introduction}

\begin{figure}[!ht]
\centering \includegraphics[width=1.00\columnwidth]{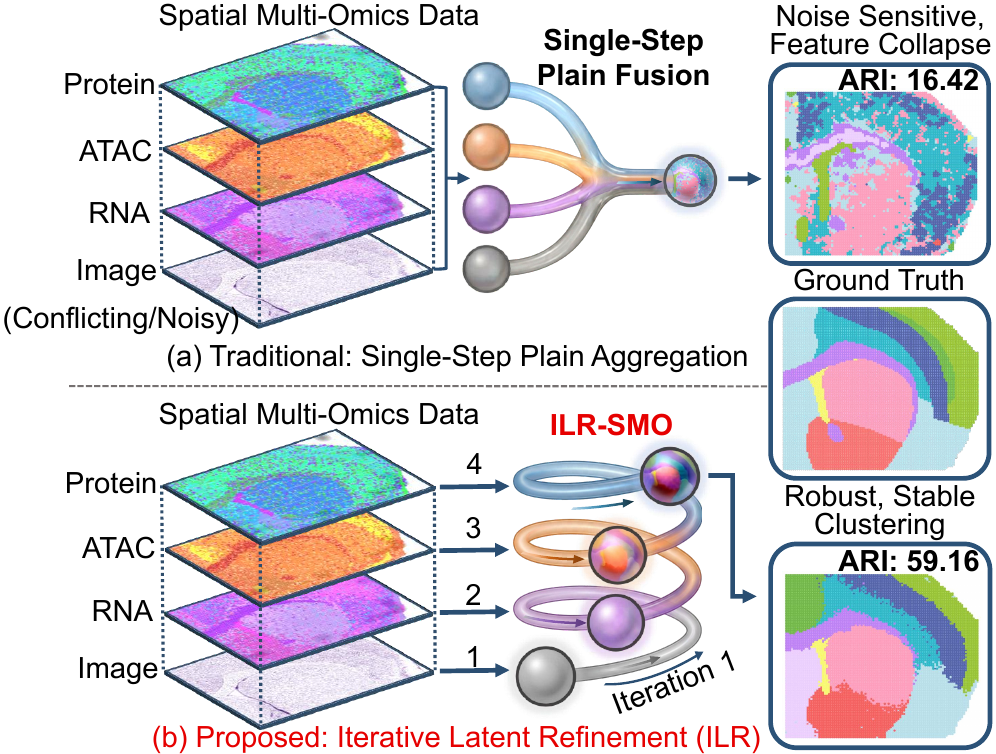}
\caption{Motivation and overview. ILR-SMO replaces single-step fusion with iterative latent refinement for robust spatial domain identification.} \label{f1} \end{figure}

Spatial multi-omics technologies enable the simultaneous measurement of multiple molecular modalities, such as RNA expression, chromatin accessibility (ATAC), protein abundance, and increasingly histological context, within the same intact tissue section. By jointly profiling complementary molecular layers with preserved spatial organization, spatial multi-omics enables a more comprehensive characterization of regulatory programs and tissue architecture than single-modality or dissociated assays~\cite{nature1,nature2}. Effective computational integration has become essential for extracting biologically meaningful insights, with broad applications across cancer biology, neuroscience, and developmental biology~\cite{deng2022spatial,zhang2023spatial,cancer}.

Existing computational approaches have made notable progress toward multimodal integration, while several key limitations remain. In practical spatial multi-omics settings, modalities exhibit pronounced heterogeneity in feature dimensionality, signal quality, and biological content, along with modality-specific noise and potentially conflicting signals. Early multi-omics methods developed for single-cell data, such as GLUE~\cite{glue}, totalVI~\cite{totalvi}, and MultiVI~\cite{multivi}, can effectively align multiple modalities but cannot exploit spatial context. More recently, SpatialGlue~\cite{spatialglue} and its extensions such as PRAGA~\cite{praga}, COSMOS~\cite{cosmos}, SWITCH~\cite{switch}, and MultiGATE~\cite{multigate}, incorporate spatial adjacency through neighborhood graphs or graph neural networks, enabling joint modeling of molecular features and spatial structure. However, most of these approaches rely on single-step plain fusion strategies, where all modalities are roughly aggregated into a shared embedding in a single step (Fig. \ref{f1}(a)). Such single-step fusion can be sensitive to measurement noise and modality heterogeneity, and often struggles to maintain stable performance across diverse datasets. In addition, most existing methods are trained end-to-end under a fixed and often bi-modal configuration, which limits extensibility as new modalities merge and makes it difficult to incorporate new data types without retraining. Furthermore, support for histopathological image modalities remains scarce. While MISO~\cite{miso} incorporates histological images into spatial multi-omics analysis, it relies on uniform pairwise cross-modal interactions, limiting its ability to prioritize informative modalities over noisy ones. These limitations highlight the need for a more robust and extensible fusion paradigm.

In this work, we propose ILR-SMO, a unified framework that performs Iterative Latent Refinement for Spatial Multi-Omics integration. Instead of treating multimodal fusion as a static aggregation process, ILR-SMO models it as a stability-aware optimization process over a shared latent representation (Fig. \ref{f1}(b)). To cope with severe modality heterogeneity and noise, the shared representation is refined progressively by integrating modality-specific information in a sequential manner. This incremental refinement allows multimodal signals to be incorporated in a controlled way, enabling later updates to correct or compensate for earlier ones and resulting in more stable optimization dynamics under weak self-supervision. To further handle uneven modality quality, ILR-SMO adopts reliability-aware modality gating, which adaptively modulates modality contributions during fusion based on their reliability. The overall learning process is guided by a stability-aware objective that jointly enforces spatial coherence and prevents representation collapse. 

Our main contributions are summarized as follows:
\begin{itemize}
    \item We propose ILR-SMO, a unified spatial multi-omics fusion framework that formulates multimodal integration as an iterative and stability-aware optimization process, supporting flexible combinations of modalities and enabling robust representation learning under modality heterogeneity and noise.
    \item We introduce iterative latent refinement with sequential latent updates and reliability-aware modality gating, together with a stability-aware joint optimization objective that explicitly controls optimization dynamics for self-supervised spatial clustering.
    \item We demonstrate the effectiveness of ILR-SMO on five spatial multi-omics benchmarks, where it consistently outperforms strong state-of-the-art baselines and exhibits robust performance across modalities{\renewcommand{\thefootnote}{\fnsymbol{footnote}}\footnotemark[2]}.
\end{itemize}

\section{Related Work}

\subsection{Spatial Multi-Omics Aggregation}

Spatial multi-omics has become an increasingly important paradigm for analyzing spatial biological data. Spatial multi-omics integrates complementary molecular signals, motivating multimodal fusion methods for spatial domain identification. Early efforts primarily focused on single-cell multi-omics integration, including  totalVI~\cite{totalvi}, GLUE~\cite{glue}, and MultiVI~\cite{multivi}, which model cross-modal relationships at the cellular level but do not incorporate spatial information. Recently, a series of spatial multi-omics methods have been proposed to jointly model molecular features and spatial adjacency, such as CellCharter~\cite{cellcharter}, SpaMultiVAE~\cite{spavae}, and SpatialGlue~\cite{spatialglue}, as well as subsequent extensions including PRAGA~\cite{praga}, COSMOS~\cite{cosmos}, SWITCH~\cite{switch}, and MultiGATE~\cite{multigate}. However, existing spatial methods commonly adopt single-step fixed fusion schemes, which limit robustness and scalability. MISO~\cite{miso} extends spatial multi-omics integration by incorporating histological images, but treats all modalities uniformly without adapting to modality reliability. Overall, existing methods advance spatial multi-omics analysis while leaving challenges in robust and flexible multimodal fusion.

\subsection{Multimodal Fusion}

Multimodal representation learning aims to integrate multiple data inputs or latent representations into a unified embedding for downstream tasks. Existing fusion strategies typically fall into static operators (e.g., concatenation~\cite{concat,switch}) and learnable mechanisms (e.g., attention-based fusion~\cite{attention1,attention2}), where modality-specific encoders first produce latent features that are then aggregated into a shared representation. However, many fusion pipelines are built upon fixed fusion structures, often assuming bi-modal or pre-defined interaction patterns~\cite{cosmos,multigate}. Such rigid fusion schemes limit the flexibility of modality interaction modeling and prevent adaptive correction when modalities exhibit imbalanced quality or noise, making the resulting representations sensitive to modality heterogeneity~\cite{chen2021multimodal,attention2}. MM-Lego~\cite{mmlego} explores frequency-domain interactions for cross-modal aggregation, highlighting the potential of more modular and extensible fusion designs. Inspired by these advances, we adapt iterative and reliability-aware fusion principles to the spatial multi-omics setting, enabling robust domain identification under heterogeneous modalities.

\section{Method}

\begin{figure*}[!ht]
\centering \includegraphics[width=1.00\textwidth]{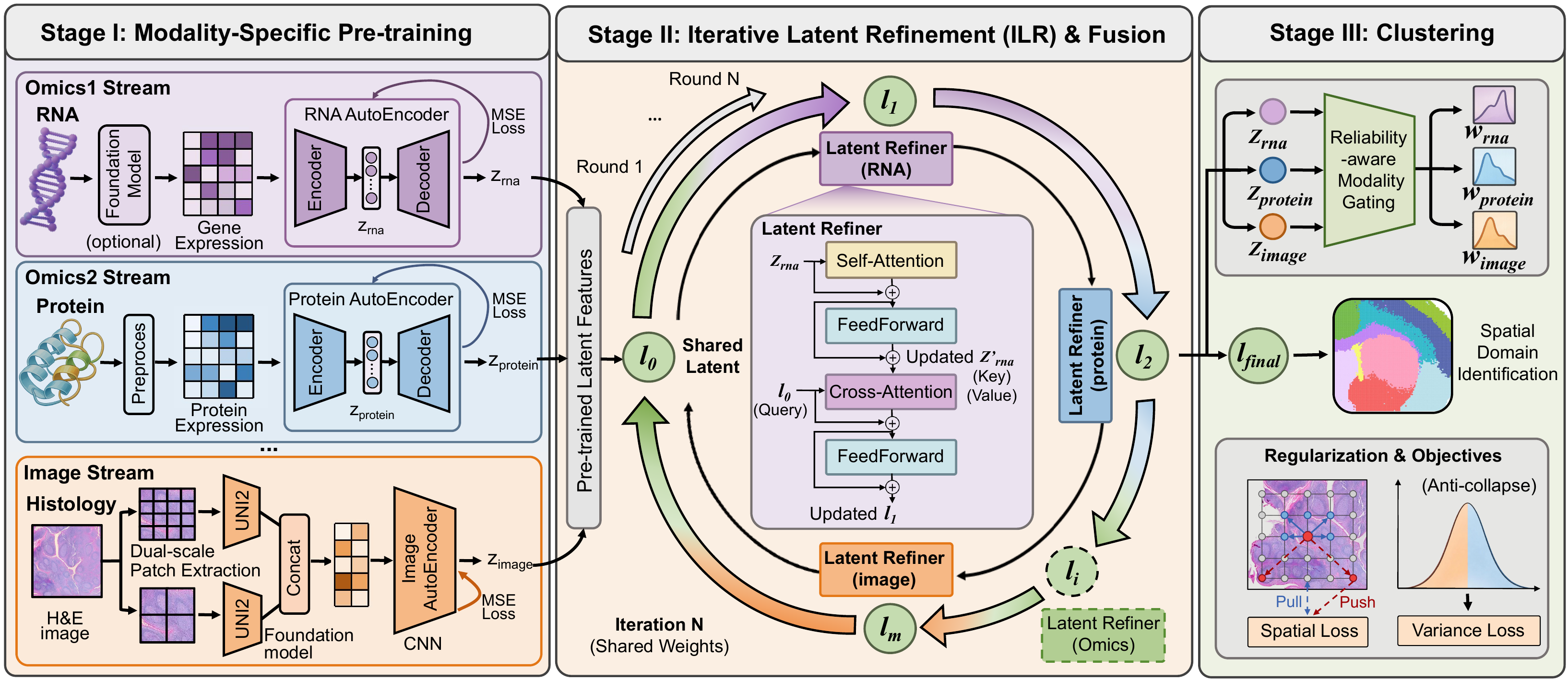}
\caption{Overview of ILR-SMO. The framework first performs modality-specific pretraining to obtain pretrained latent embeddings, then applies Iterative Latent Refinement (ILR) via sequential latent updates to progressively integrate heterogeneous modalities into a shared representation. Reliability-aware modality gating adaptively modulates refinement strength, while spatial consistency and dispersion regularization ensure stable, coherent, and non-collapsing optimization, yielding a unified latent for spatial domain identification.} \label{f2} \end{figure*}

We propose ILR-SMO, a unified framework for spatial multi-omics integration that learns a shared latent representation for unsupervised spatial domain identification. As shown in Fig. \ref{f2}, ILR-SMO comprises modality-specific pretraining, iterative multi-modal integration, and clustering-based spatial domain identification.

\subsection{Preliminaries}

Given a spatial multi-omics dataset consisting of $N$ spatial locations measured across $M$ omics and imaging streams, we denote the preprocessed modality-specific input matrix of the $m$-th omics stream as $X^{(m)} \in \mathbb{R}^{N \times d_m}$, where $d_m$ is the input dimensionality of modality $m$, and $\mathbf{x}^{(m)}_i \in \mathbb{R}^{d_m}$ is the input at location $i$. Each location is associated with a spatial coordinate $\mathbf{s}_i \in \mathbb{R}^{2}$, collected in $S \in \mathbb{R}^{N \times 2}$. Each data stream is encoded to a modality-specific latent space via%
\begin{align}
\mathbf{z}^{(m)}_i = f_m\left(\mathbf{x}^{(m)}_i\right), \quad \mathbf{z}^{(m)}_i \in \mathbb{R}^{d_z}.
\end{align}%
Given the latent representations $\{\mathbf{z}_i^{(m)}\}_{m=1}^M$, a shared latent representation \( \mathbf{l}_i \) is constructed for each spatial location as%
\begin{align}
\mathbf{l}_i = g\!\left( \mathbf{z}^{(1)}_i, \ldots, \mathbf{z}^{(M)}_i \right),
\end{align}%
which serves as the unified embedding for unsupervised spatial domain identification using a clustering algorithm \( \mathcal{C}(\cdot) \), such as K-means or Leiden.

\subsection{Modality-Specific Pretraining}
As shown in Fig. \ref{f2} (stage I), we independently pretrain an autoencoder for each modality to obtain compact latent representations that serve as inputs for subsequent cross-modal integration. The inputs to each modality encoder are modality-specific feature vectors obtained after standard preprocessing or foundation-model-based feature extraction. In particular, image modality is represented by dual-scale patch features extracted around each spatial location and encoded using a pathology foundation model (UNI2~\cite{uni}), whose outputs are concatenated to capture both local and contextual visual patterns. Given the preprocessed input $\mathbf{x}_i^{(m)}$, each modality $m$ is associated with an encoder–decoder pair $(f_m^{enc}, f_m^{dec})$, which maps the input to a latent embedding and reconstructs it as:%
\begin{align}
\mathbf{z}_i^{(m)} = f_m^{enc}(\mathbf{x}_i^{(m)}), \quad
\hat{\mathbf{x}}_i^{(m)} = f_m^{dec}(\mathbf{z}_i^{(m)}).
\end{align}%
The autoencoders are trained independently by minimizing a mean squared reconstruction error (MSE),%
\begin{align}
\mathcal{L}_{\mathrm{rec}}^{(m)}=\frac{1}{N}\sum_{i=1}^N\left\|\hat{\mathbf{x}}_i^{(m)}-\mathbf{x}_i^{(m)}\right\|_2^2,
\end{align}%
and the total stage I objective is the sum of reconstruction losses across all modalities. This pretraining decouples modality-specific representation learning from cross-modal fusion, providing informative and comparable latent embeddings for the iterative refinement stage. 

\subsection{Iterative Latent Refinement}

\subsubsection{Optimization Perspective of ILR} As illustrated in Fig. \ref{f2} (stage II), we view multi-modal fusion as an optimization process over a shared latent representation $l$, driven by pretrained modality-specific latents that provide heterogeneous and potentially noisy supervision. In spatial multi-omics settings, uneven modality quality and view-specific biases make single-step aggregation unstable under unsupervised learning. Iterative Latent Refinement (ILR) addresses this by sequentially integrating modality information into the shared latent, injecting one modality at a time through incremental updates. This formulation allows each modality to contribute in a controlled manner while enabling subsequent updates to compensate for earlier deviations, resulting in smoother optimization dynamics and improved robustness for self-supervised spatial domain identification.

\subsubsection{Mechanism of Sequential Latent Updates} ILR is instantiated by an Iterative Latent Refinement Module (ILRM), which maintains a learnable shared latent representation $L$ and updates it via Sequential Latent Updates (SLU) across modalities. Given pretrained modality-specific latent embeddings from stage I, where $Z^{(m)} \in \mathbb{R}^{B \times 1 \times D}$ denotes the normalized encoder output, ILRM initializes a shared latent token $L \in \mathbb{R}^{B \times 1 \times D}$ and refines it over multiple refinement rounds. In each round, SLU traverses modalities in a fixed order and applies a Modality-wise Latent Refiner (MLR) to sequentially inject modality context into the shared latent. For a modality $m$, MLR first updates the modality token using a pre-norm self-attention module,%
\begin{align}
Z^{(m)\,*} = Z^{(m)} + \mathrm{SelfAttention}\!\big(\mathrm{LayerNorm}(Z^{(m)})\big),
\end{align}%
\begin{align}
Z^{(m)\,'} = Z^{(m)\,*} + \mathrm{FeedForward}\!\big(\mathrm{LayerNorm}(Z^{(m)\,*})\big).
\end{align}%
The updated modality token $Z^{(m)'}$ is then used as context to refine the shared latent via cross-attention,%
\begin{equation}
    \resizebox{.91\linewidth}{!}{$
            \tilde L = L +\mathrm{CrossAttention}\big(\mathrm{LayerNorm}(L),\mathrm{LayerNorm}(Z^{(m)'})\big),
        $}
\end{equation}%
\begin{align}
L_{updated} = \tilde L +\mathrm{FeedForward}\big(\mathrm{LayerNorm}(\tilde L)\big).
\end{align}%
The refined latent $L_{updated}$ is passed to the next MLR, and one SLU round is completed after all modalities are processed. Repeating this procedure for multiple rounds yields the final shared representation $L_{final}$. Compared to single-step plain fusion schemes, SLU realizes ILR as a sequence of modality-conditioned refinements, providing a structured mechanism for cross-modal integration.

\subsubsection{Stability Analysis of ILR} To analyze the stability of Iterative Latent Refinement, we adopt an optimization-level abstraction by modeling the refinement of the shared latent $L$ as minimizing a composite objective%
\begin{align}
F(L)=\sum_{m=1}^{M} f_m(L) + r(L),
\end{align}%
where each $f_m$ represents a modality-specific contribution and $r(L)$ collects regularization terms. Let $g_m(L)=\nabla f_m(L)$ and assume each $f_m$ has $\beta$-Lipschitz continuous gradients. We compare ILR implemented via Sequential Latent Updates (SLU) with a single-step joint latent update that aggregates all modality gradients at once. 

(i) First-order equivalence with second-order smoothing. A joint update takes the form,
\begin{align}
\Delta_{\text{joint}} = -\eta \sum_{m=1}^{M} g_m(L),
\end{align}%
while one SLU cycle yields:
\begin{align}
\Delta_{\text{SLU}} = -\eta \sum_{m=1}^{M} g_m(L) + O(\eta^2),
\end{align}%
with the deviation bounded by
\begin{align}
\|\Delta_{\text{SLU}}-\Delta_{\text{joint}}\|
\le \beta\eta^2(M-1)\sum_{m=1}^{M}\|g_m(L)\|.
\end{align}%
This shows that SLU preserves the dominant descent direction while introducing higher-order corrections that smooth the optimization trajectory.

(ii) Localized impact of noisy modalities. For a perturbed modality $k$ with $\tilde g_k(L)=g_k(L)+\epsilon(L)$, $\|\epsilon(L)\|\le E$. 
$L^{+}_{\text{SLU}}$ and $\tilde L^{+}_{\text{SLU}}$ denote the SLU outputs under clean and perturbed gradients, respectively. The deviation satisfies
\begin{align}
\|\tilde L^{+}_{\text{SLU}} - L^{+}_{\text{SLU}}\|
\le \eta E e^{\eta\beta M}
\approx O(\eta E),
\end{align}%
for typical small step sizes, indicating that modality-specific noise affects only a bounded incremental update instead of dominating the entire refinement.

(iii) Alleviating gradient conflicts. For joint aggregation, the squared norm of the summed gradient expands as
\begin{align}
\|\sum_{m=1}^{M} g_m(L)\|^2
= \sum_{m=1}^{M}\|g_m(L)\|^2
- 2\sum_{i<j} g_i(L)^\top g_j(L), 
\end{align}%
where negative inner products cause cancellation under conflicting modalities. SLU avoids this hard aggregation by applying gradients sequentially, effectively decomposing refinement into modality-conditioned sub-steps and yielding more consistent optimization under heterogeneous modalities. 

This abstraction emphasizes update order and locality rather than architectural details, and serves as an approximation of the ILR mechanism. Complete derivations and additional bounds are provided in Appendix A.

\subsection{Reliability-Aware Modality Gating}
\label{sec3.4}
To account for heterogeneous modality quality in spatial multi-omics data, we introduce Reliability-aware Modality Gating (RMG) to adaptively modulate refinement strength at the sample and modality levels, as shown in Fig. \ref{f2} (stage III). RMG takes normalized modality-specific latent embeddings ${\{\mathbf{z}^{(m)}_i}\}_{m=1}^{M}$ as input and produces a soft reliability weight vector $\mathbf{w}_i \in \mathbb{R}^M$. Rather than modifying the fusion mechanism, RMG controls the update magnitude by weighting the alignment between the fused latent and each modality,
\begin{align}
\mathcal{L}_{\mathrm{align}}= \frac{1}{N}\sum_{i=1}^{N}\sum_{m=1}^{M}w_{i,m}\|\mathbf{z}^{f}_i-\mathbf{z}^{(m)}_i\|_2^2,
\end{align}%
where $\mathbf{z}^{f}i$ denotes the fused representation. To avoid degenerate solutions where the gating collapses onto a single modality, we regularize the batch-averaged weights $\bar{\mathbf{w}}$ using a negative entropy term, 
\begin{align}
\mathcal{L}_{\mathrm{ent}}= \sum_{m=1}^{M} \bar{w}_m \log(\bar{w}_m+\epsilon).
\end{align}%
Together with ILR, RMG adaptively amplifies informative modalities while suppressing unreliable ones, improving robustness under modality heterogeneity and noise.

\begin{table*}[ht]
\centering
\renewcommand{\arraystretch}{1.2}
\setlength{\tabcolsep}{1.5pt}
\resizebox{\textwidth}{!}{%
\begin{tabular}{l|rrrrrr|rrrrrr}
\hline
\multicolumn{1}{c}{} & \multicolumn{6}{c}{Mouse Brain Dataset} & \multicolumn{6}{c}{Human Lymph Node Dataset} \\
\hline
\textbf{Methods} & \textbf{ARI(\%)} & \textbf{Homo.(\%)} & \textbf{MI(\%)} & \textbf{V-meas.(\%)} & \textbf{AMI(\%)} & \textbf{NMI(\%)} & \textbf{ARI(\%)} & \textbf{Homo.(\%)} & \textbf{MI(\%)} & \textbf{V-meas.(\%)} & \textbf{AMI(\%)} & \textbf{NMI(\%)} \\
\hline
MultiVI & - & - & - & - & - & - & 7.78 & 8.18 & 13.79 & 8.74 & 8.32 & 8.74 \\
SpatialGlue & \underline{35.45} & 49.34 & 98.59 & 48.20 & 48.10 & 48.20 & 26.98 & 39.34 & 66.37 & 38.48 & 38.19 & 38.48 \\
MISO & 22.52 & 31.95 & 63.84 & 33.52 & 33.38 & 33.52 & 24.28 & 28.09 & 47.39 & 35.49 & 35.18 & 35.49 \\
PRAGA & 31.91 & 47.33 & 94.57 & 46.90 & 46.80 & 46.90 & \underline{28.23} & \underline{40.23} & \underline{67.87} & \underline{39.08} & \underline{38.80} & \underline{39.08} \\
COSMOS & 34.84 & \underline{61.94} & \underline{123.76} & \underline{59.88} & \underline{59.80} & \underline{59.88} & 18.65 & 33.34 & 56.25 & 31.18 & 30.87 & 31.18 \\
SWITCH & 16.42 & 28.43 & 56.81 & 30.45 & 30.29 & 30.45 & 21.08 & 28.46 & 48.02 & 30.63 & 30.29 & 30.63 \\
MultiGATE & 31.62 & 52.12 & 104.15 & 48.88 & 48.78 & 48.88 & 18.18 & 36.00 & 60.74 & 34.20 & 33.90 & 34.20 \\
ILR-SMO & \textbf{59.16} & \textbf{73.26} & \textbf{146.38} & \textbf{69.97} & \textbf{69.91} & \textbf{69.97} & \textbf{28.38} & \textbf{40.88} & \textbf{68.96} & \textbf{39.11} & \textbf{38.83} & \textbf{39.11} \\
\hline
\multicolumn{1}{c}{} & \multicolumn{6}{c}{Human Hippocampus Dataset} & \multicolumn{6}{c}{Simulated Mouse Visual Cortex Dataset} \\
\hline
\textbf{Methods} & \textbf{ARI(\%)} & \textbf{Homo.(\%)} & \textbf{MI(\%)} & \textbf{V-meas.(\%)} & \textbf{AMI(\%)} & \textbf{NMI(\%)} & \textbf{ARI(\%)} & \textbf{Homo.(\%)} & \textbf{MI(\%)} & \textbf{V-meas.(\%)} & \textbf{AMI(\%)} & \textbf{NMI(\%)} \\
\hline
MultiVI & 15.17 & 28.83 & 50.11 & 27.61 & 27.32 & 27.61 & 10.30 & 14.64 & 24.90 & 14.58 & 14.06 & 14.58 \\
SpatialGlue & 32.82 & 44.24 & 75.69 & 41.85 & 41.58 & 41.85 & 60.58 & 64.20 & 109.18 & 63.11 & 62.88 & 63.11 \\
MISO & 19.62 & 30.32 & 52.70 & 30.39 & 30.09 & 30.39 & 19.58 & 24.77 & 42.12 & 24.51 & 24.04 & 24.51 \\
PRAGA & 17.06 & 29.05 & 49.69 & 27.45 & 27.12 & 27.45 & 34.24 & 41.95 & 71.33 & 40.92 & 40.56 & 40.92 \\
COSMOS & 31.37 & 47.87 & 83.22 & 47.00 & 46.78 & 47.00 & \underline{73.58} & \underline{79.65} & \underline{135.45} & \underline{78.05} & \underline{77.92} & \underline{78.05} \\
SWITCH & 18.31 & 27.82 & 48.36 & 30.98 & 30.65 & 30.98 & 21.14 & 34.76 & 59.12 & 36.75 & 36.32 & 36.75 \\
MultiGATE & \underline{45.96} & \underline{49.16} & \underline{85.45} & \underline{49.80} & \underline{49.58} & \underline{49.80} & 66.65 & 75.44 & 128.30 & 74.48 & 74.32 & 74.48 \\
ILR-SMO & \textbf{53.40} & \textbf{54.39} & \textbf{94.54} & \textbf{54.25} & \textbf{54.06} & \textbf{54.25} & \textbf{74.20} & \textbf{79.82} & \textbf{135.74} & \textbf{79.12} & \textbf{78.99} & \textbf{79.12} \\
\hline
\end{tabular}
}
\caption{Quantitative results on the mouse brain dataset, the human lymph node dataset, the human hippocampus dataset, and the simulated mouse visual cortex dataset. \textbf{Best} and \underline{Second} results are in highlight. MultiVI is omitted on the Mouse Brain dataset because only preprocessed data are available, while MultiVI requires raw count–based inputs. V-measure and NMI calculated with arithmetic-mean normalization are mathematically equivalent and therefore yield identical values.}
\label{tab:main}
\end{table*}

\subsection{Stability-Aware Joint Optimization}

We adopt a Stability-aware Joint Optimization (SJO) objective to guide ILR toward spatially coherent and non-collapsing solutions. Spatial Consistency Regularization promotes local smoothness by pulling together spatial neighbors while separating random negatives,
\begin{align}
\mathcal{L}_{\mathrm{spatial}}
= \frac{1}{N}\sum_{i=1}^N \Bigg(
& \frac{1}{|\mathcal{N}_i|}\sum_{j\in\mathcal{N}_i}
\|\mathbf{z}^f_i-\mathbf{z}^f_j\|_2 \nonumber \\
& + \max\!\left(0,\;
\gamma-\|\mathbf{z}^f_i-\mathbf{z}^f_{\pi(i)}\|_2
\right)
\Bigg).
\end{align}%
while representation dispersion regularization prevents collapse by enforcing a minimum per-dimension variance,
\begin{align}
\mathcal{L}_{\mathrm{var}}=\frac{1}{d}\sum_{k=1}^{d}\max\left(0,\sigma_0-\sqrt{\mathrm{Var}(z^f_{\cdot,k})+\epsilon}\right).
\end{align}%
Cross-modal consistency is encouraged via pairwise agreement among modality latents,
\begin{align}
\mathcal{L}_{\mathrm{contr}}=\sum_{1\le m<n\le M}\frac{1}{N}\sum_{i=1}^N\|\mathbf{z}^{(m)}_i-\mathbf{z}^{(n)}_i\|_2^2,
\end{align}%
while reliability-weighted alignment and gating entropy are defined in Sec. \ref{sec3.4}. The overall objective is
\begin{align}
\mathcal{L}_{total}=\alpha_{rec}\mathcal{L}_{rec}+\alpha_{contr}\mathcal{L}_{\mathrm{contr}}+\alpha_{align}\mathcal{L}_{\mathrm{align}} \nonumber \\ +\alpha_{spatial}\mathcal{L}_{\mathrm{spatial}}+\alpha_{var}\mathcal{L}_{\mathrm{var}}+\alpha_{ent}\mathcal{L}_{\mathrm{ent}},
\end{align}%
with coefficients $\alpha_*$. Together, these terms provide directional guidance and anti-collapse constraints, enabling stable refinement under heterogeneous multi-modal noise.

\section{Experiments}

\begin{table}[t]
\centering
\renewcommand{\arraystretch}{1.2}
\setlength{\tabcolsep}{1.2pt}
\resizebox{\columnwidth}{!}{%
\begin{tabular}{l|rrrrrr}
\hline
\multicolumn{1}{c}{} & \multicolumn{6}{c}{Spatial Simulated Triplet Omics Dataset} \\
\hline
\textbf{Methods} & \textbf{ARI(\%)} & \textbf{Homo.(\%)} & \textbf{MI(\%)} & \textbf{V-meas.(\%)} & \textbf{AMI(\%)} & \textbf{NMI(\%)} \\
\hline
MultiVI & 34.54 & 53.07 & 82.12 & 48.91 & 48.63 & 48.91 \\
SpatialGlue & 97.25 & 96.55 & 149.40 & 96.55 & 96.54 & 96.55 \\
MISO & 15.13 & 24.26 & 37.54 & 28.97 & 28.63 & 28.97 \\
PRAGA & \underline{99.37} & \underline{99.10} & \underline{153.45} & \underline{99.07} & \underline{99.07} & \underline{99.07} \\
ILR-SMO & \textbf{100.00} & \textbf{100.00} & \textbf{154.74} & \textbf{100.00} & \textbf{100.00} & \textbf{100.00} \\
\hline
\end{tabular}
}
\caption{Quantitative results on the spatial simulated triplet-omics dataset using six evaluation metrics. COSMOS, Switch, and MultiGATE are omitted as they support only bimodal inputs.}
\label{tab:triplet}
\end{table}

\subsection{Experimental Setups}

\begin{figure*}[!ht]
\centering \includegraphics[width=0.96\textwidth]{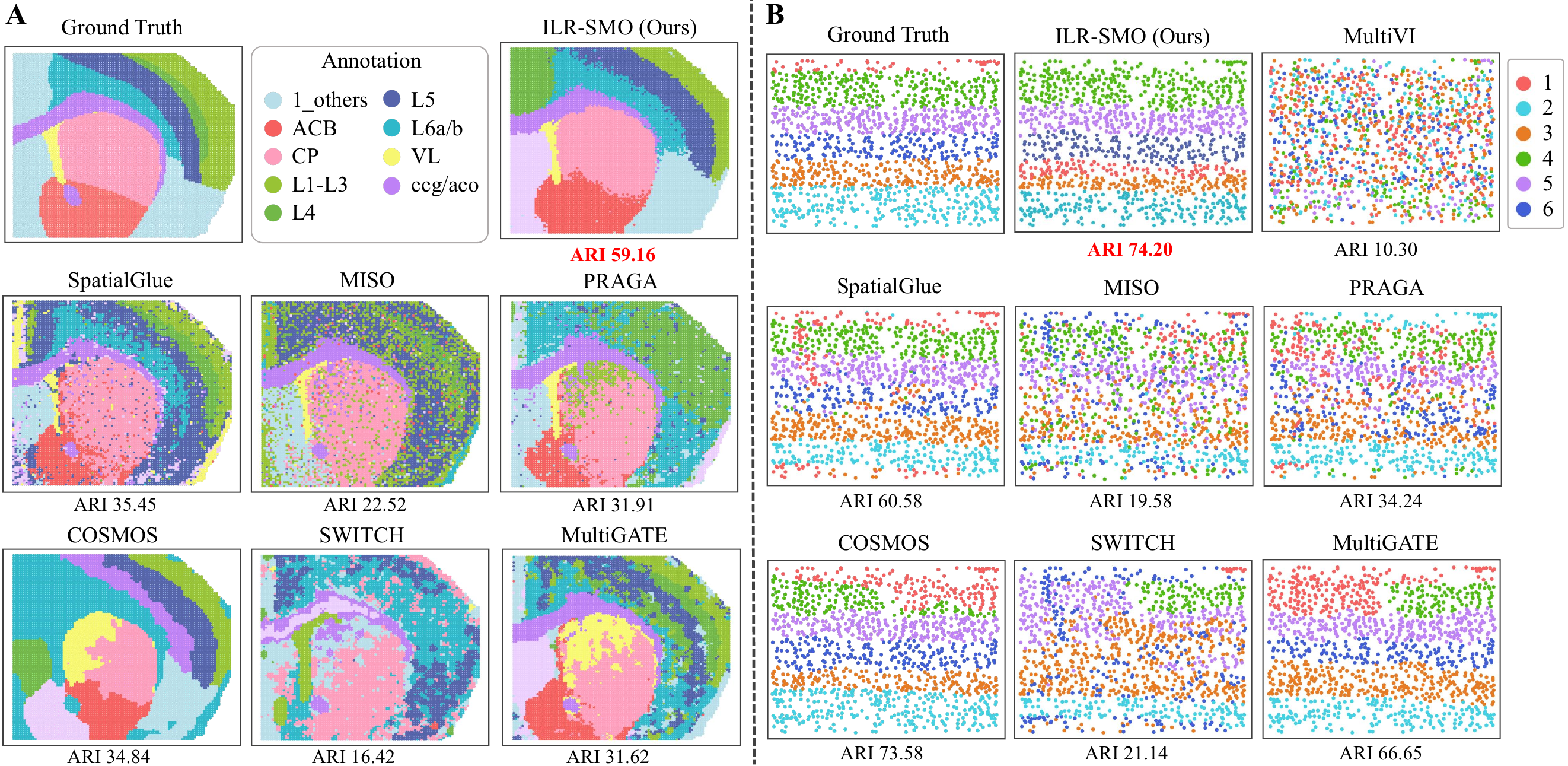}
\caption{A: Ground truth annotations of brain regions in the mouse brain dataset and spatial domain identification results for various methods. B: Ground truth annotations of the mouse visual cortex dataset and spatial domain identification results for various methods.} \label{f3} \end{figure*}

\paragraph{Datasets.} We evaluate our proposed method on five spatial multi-omics benchmarks with diverse modalities: (1) Spatial epigenome–transcriptome mouse brain dataset~\cite{cosmos}; (2) Human Lymph Node dataset~\cite{spatialglue}; (3) Human Hippocampus dataset~\cite{multigate}; (4) Simulated Mouse Visual Cortex dataset~\cite{cosmos}; (5) Spatial simulated triplet omics dataset~\cite{spatialglue}. Details are provided in Appendix B.

\paragraph{Baseline Methods.} We compare ILR-SMO against a set of state-of-the-art multi-omics methods. We include MultiVI~\cite{multivi} as a representative multi-modal omics integration method, along with six latest spatial multi-omics approaches: SpatialGlue~\cite{spatialglue}, MISO~\cite{miso}, PRAGA~\cite{praga}, COSMOS~\cite{cosmos}, SWITCH~\cite{switch}, and MultiGATE~\cite{multigate}. Among these baselines, MISO is the only method that incorporates image features.

\paragraph{Evaluation Metrics.} We adopt six widely used clustering metrics: Adjusted Rand Index (ARI), Homogeneity (Homo.), Mutual Information (MI), V-measure (V-meas.), Adjusted Mutual Information (AMI), and Normalized Mutual Information (NMI) ~\cite{spatialglue}, which provide complementary perspectives on clustering quality. 

\paragraph{Implementation Details.} The entire model is implemented using PyTorch and experiments are conducted on a single NVIDIA GeForce RTX 3090 GPU. 
Despite slightly increased parameterization, ILR-SMO maintains moderate computational cost and is efficiently trainable on standard consumer-grade hardware, with detailed analyses in Appendix G.

\subsection{Experimental Results}

\paragraph{Quantitative Comparison.} To comprehensively evaluate the effectiveness of ILR-SMO, we report the quantitative clustering results on five spatial multi-omics datasets using six widely adopted evaluation metrics (Tables \ref{tab:main} and \ref{tab:triplet}). ILR-SMO consistently achieves state-of-the-art performance across all datasets and metrics, demonstrating strong and robust spatial domain identification capability under diverse experimental settings. Notably, ILR-SMO exhibits substantial performance margins over competing methods. On the mouse brain dataset, ILR-SMO surpasses the second-best result by 23.71\% in ARI and 22.62\% in MI, while on the human hippocampus dataset it outperforms the strongest baseline MultiGATE by 7.44\% in ARI and 9.09\% in MI. 
ILR-SMO further maintains superior performance on the simulated mouse visual cortex dataset with structured perturbations and noise, indicating strong robustness against external disturbances. In contrast, competing methods exhibit larger variability across datasets.
Finally, ILR-SMO naturally extends to tri-modal integration and achieves the best performance across all metrics on the spatial simulated triplet omics dataset (Table \ref{tab:triplet}), demonstrating strong scalability across modalities.

\paragraph{Qualitative Visualization.}

We visualize the spatial domain identification results on the mouse brain dataset and the simulated mouse visual cortex dataset. As shown in Fig. \ref{f3}, ILR-SMO produces spatial patterns highly consistent with the ground-truth annotations. On the mouse brain dataset, ILR-SMO yields compact spatial domains with clear boundaries and fewer scattered regions, leading to more coherent spatial organization. Notably, ILR-SMO is the only method that correctly distinguishes the nucleus accumbens (ACB) and caudoputamen (CP) regions. On the simulated mouse visual cortex dataset, ILR-SMO also recovers well-defined laminar structures that closely match the ground truth, demonstrating its ability to capture hierarchical spatial organization.

\subsection{Robustness Under Noisy Modalities}

\begin{table*}[t]
\centering
\renewcommand{\arraystretch}{1.2}
\setlength{\tabcolsep}{1.2pt}
\resizebox{\textwidth}{!}{%
\begin{tabular}{l|rrrrrr|rrrrrr}
\hline
\multicolumn{1}{c}{} & \multicolumn{6}{c}{Mouse Brain Dataset} & \multicolumn{6}{c}{Simulated Mouse Visual Cortex Dataset} \\
\hline
\textbf{Methods} & \textbf{ARI(\%)} & \textbf{Homo.(\%)} & \textbf{MI(\%)} & \textbf{V-meas.(\%)} & \textbf{AMI(\%)} & \textbf{NMI(\%)} &
\textbf{ARI(\%)} & \textbf{Homo.(\%)} & \textbf{MI(\%)} & \textbf{V-meas.(\%)} & \textbf{AMI(\%)} & \textbf{NMI(\%)} \\
\hline
w/o Gating & 55.39 & 69.15 & 138.16 & 66.53 & 66.46 & 66.53 & 71.74 & 77.72 & 132.16 & 76.90 & 76.76 & 76.90 \\
w/o $L_{\text{spatial}}$ & 23.10 & 35.32 & 70.57 & 33.59 & 33.47 & 33.59 & 16.02 & 21.81 & 37.10 & 21.61 & 21.12 & 21.61 \\
w/o ILR & 34.32 & 44.93 & 89.77 & 43.76 & 43.65 & 43.76 & 44.81 & 54.91 & 93.37 & 53.93 & 53.65 & 53.93 \\
w/o SLU & 56.58 & 70.70 & 141.26 & 67.56 & 67.50 & 67.56 & 68.47 & 74.85 & 127.29 & 73.23 & 73.06 & 73.23 \\
\hline
ILR-SMO & \textbf{59.16} & \textbf{73.26} & \textbf{146.38} & \textbf{69.97} & \textbf{69.91} & \textbf{69.97} &
\textbf{74.20} & \textbf{79.82} & \textbf{135.74} & \textbf{79.12} & \textbf{78.99} & \textbf{79.12} \\
\hline
\end{tabular}
}
\caption{Ablation results on the mouse brain and simulated mouse visual cortex datasets; ``w/o'' denotes ``without.''}
\label{tab:ablation}
\end{table*}

\begin{figure}[!ht]
\centering \includegraphics[width=1.00\columnwidth]{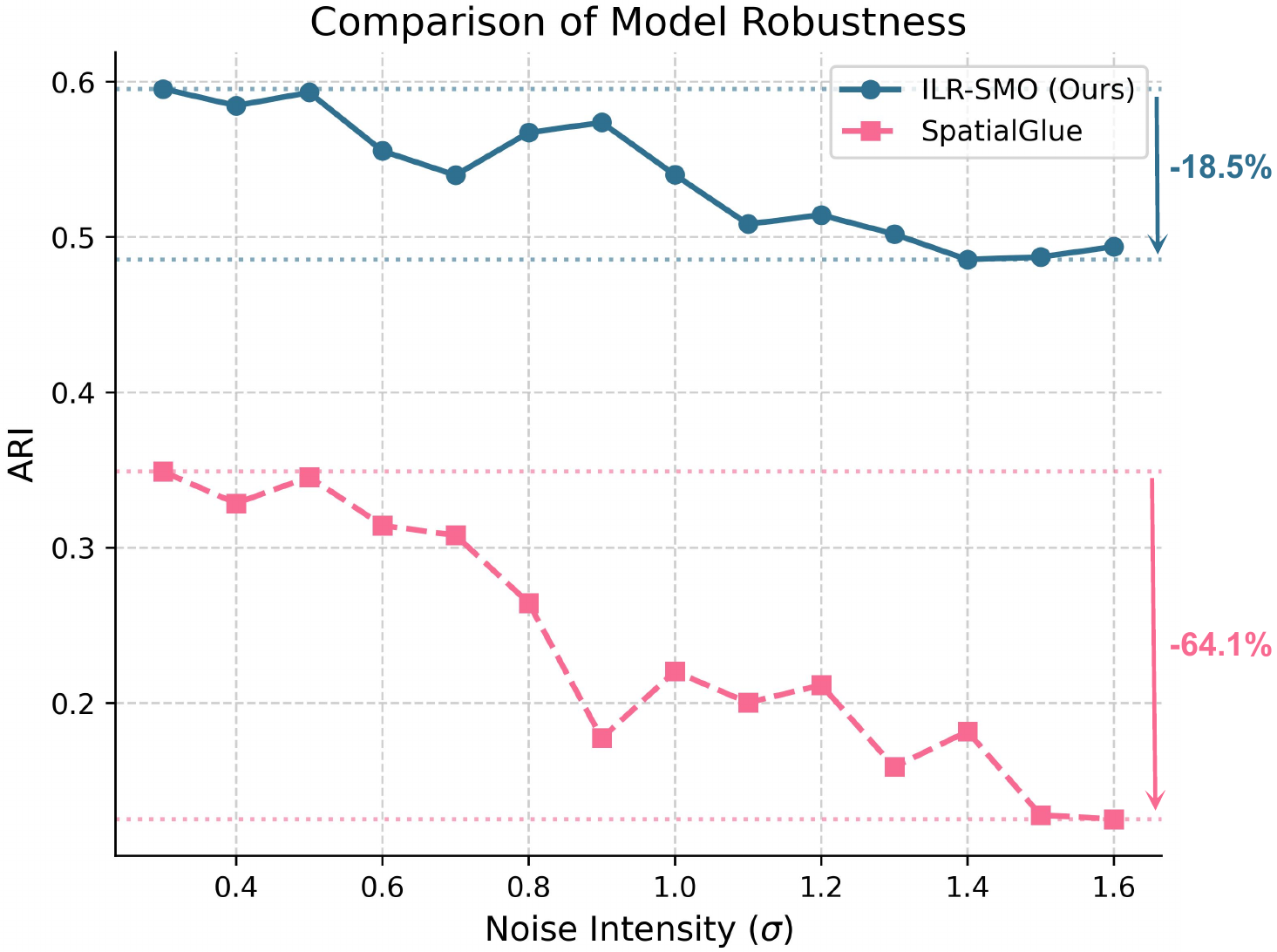}
\caption{Robustness comparison under increasing Gaussian noise injected into the ATAC modality on the mouse brain dataset.} \label{f4} \end{figure}

We further evaluate the robustness of ILR-SMO under modality corruption by injecting Gaussian noise with varying intensity $\sigma$ into the ATAC modality of the mouse brain dataset. This setting simulates realistic scenarios where a single modality is severely degraded and examines whether the model can maintain stable spatial domain identification. As shown in Fig. \ref{f4}, ILR-SMO exhibits strong robustness against increasing noise levels. Even when the noise intensity reaches $\sigma$=1.6, ILR-SMO maintains an ARI around 0.5, indicating that meaningful spatial structures are still preserved. In contrast, the baseline SpatialGlue is highly sensitive to the corrupted modality, with ARI rapidly decreasing to 0.12, corresponding to a performance drop of 64.1\% from its best result, whereas ILR-SMO only experiences a relative decrease of 18.5\%. These results demonstrate that ILR-SMO degrades gracefully under severe modality noise and is substantially more robust to modality corruption.

\subsection{Extension With Foundation Models}

\begin{figure}[!ht]
\centering \includegraphics[width=0.90\columnwidth]{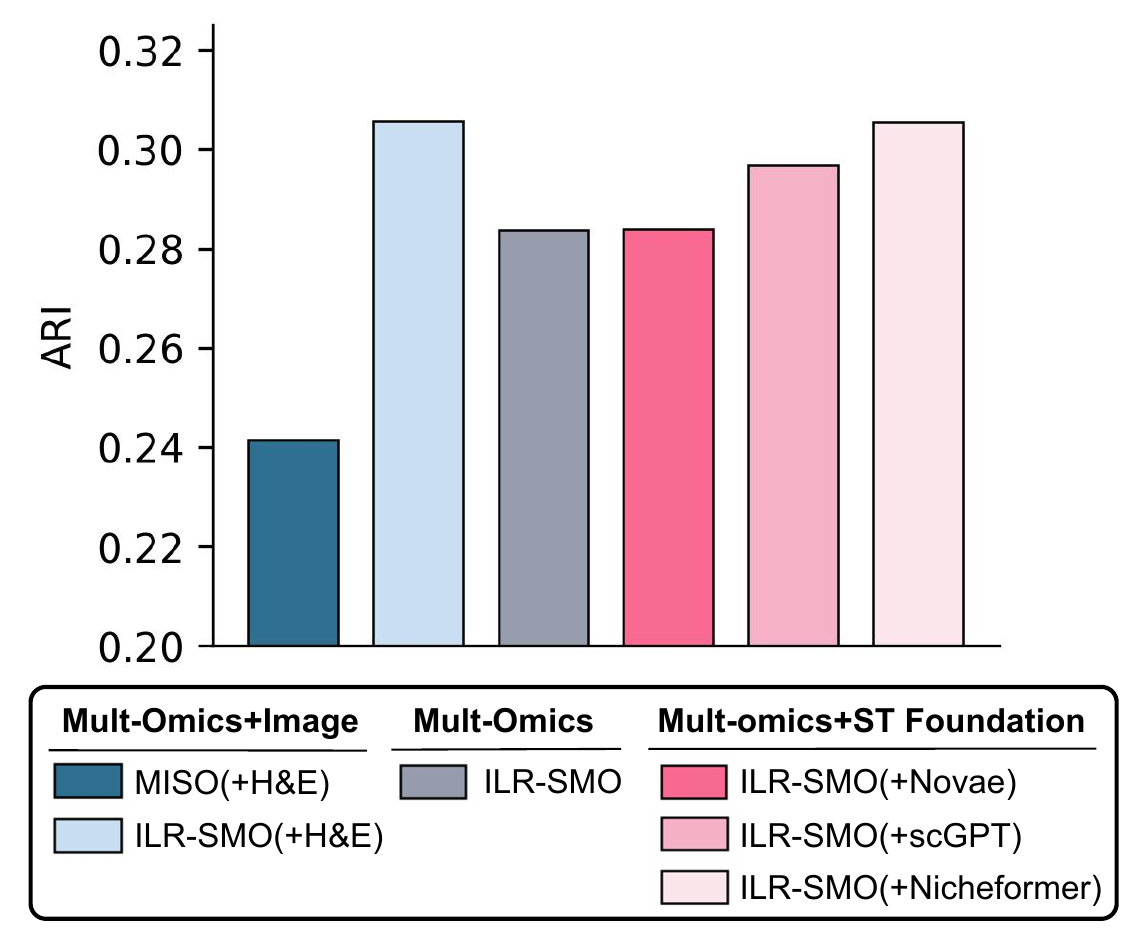}
\caption{ILR-SMO benefits from both histopathology and transcriptomic foundation representations, achieving consistent improvements under their respective modality settings.} \label{f5} \end{figure}

ILR-SMO is naturally compatible with emerging foundation models and can seamlessly incorporate their representations to further improve performance (Fig. \ref{f5}). Incorporating histopathological foundation model features leads to consistent performance gains over the original image setting and substantially outperforms MISO with the image modality. In addition, ILR-SMO can flexibly integrate transcriptomic foundation models, including Novae~\cite{novae}, scGPT~\cite{scgpt}, and Nicheformer~\cite{nicheformer}, all of which yield improvements. These results demonstrate that ILR-SMO serves as a general and extensible framework that can readily benefit from advances in foundation models across modalities.

\subsection{Ablation Studies}

We conduct ablation studies on a real-world dataset (mouse brain) and a simulated dataset (simulated mouse visual cortex) to assess the effectiveness of the key components in ILR-SMO. Specifically, we remove reliability-aware modality gating (denoted as w/o Gating), spatial consistency regularization (denoted as w/o $L_{spatial}$), replace iterative latent refinement–based fusion with concatenation followed by projection (w/o ILR), or collapse sequential latent updates into a single-step update using all modalities simultaneously (w/o SLU). As shown in Table \ref{tab:ablation}, removing any component consistently degrades performance across all metrics on both datasets, indicating that each module contributes meaningfully to the overall model. In particular, w/o ILR and w/o SLU cause pronounced drops, underscoring the importance of iterative latent refinement. Moreover, ILR is insensitive to modality update order, as all six permutations achieve identical clustering performance on the triplet-omics dataset (Appendix F). The full ILR-SMO model achieves the best performance.

\section{Conclusion}
In this work, we present ILR-SMO, a unified framework for spatial multi-omics integration that formulates multi-modal fusion as iterative optimization over a shared latent representation. By introducing Iterative Latent Refinement with sequential latent updates and Reliability-aware Modality Gating, ILR-SMO robustly integrates heterogeneous modalities under noise, imbalance, and gradient conflicts, while a stability-aware objective enforces spatial coherence and prevents representational collapse in an unsupervised setting. Extensive experiments on five diverse spatial multi-omics datasets demonstrate that ILR-SMO consistently outperforms state-of-the-art methods in spatial domain identification across multiple metrics and experimental scenarios.

\section*{Acknowledgments}
This work was supported by the Shanghai Municipal Special Program for Basic Research on General AI Foundation Models (Grant No. 2025SHZDZX025G11) and NSFC (grant nos. 62402120 to C.M., 62472102 to B.Y., and 62372117 to W.T.). The computations in this research were performed using the CFFF platform of Fudan University.

\bibliographystyle{named}
\bibliography{main}

\end{document}